%% file: main.tex
\documentclass[letterpaper]{article} 
\usepackage{aaai2026}  
\usepackage{times}  
\usepackage{helvet}  
\usepackage{courier}  
\usepackage[hyphens]{url}  
\usepackage{graphicx} 
\usepackage{natbib}  
\usepackage{caption} 
\usepackage{algorithm}
\usepackage{algorithmic}

\usepackage{newfloat}
\usepackage{listings}
\DeclareCaptionStyle{ruled}{labelfont=normalfont,labelsep=colon,strut=off} 
\floatstyle{ruled}
\newfloat{listing}{tb}{lst}{}
\floatname{listing}{Listing}
\usepackage[accsupp]{axessibility}
\usepackage{graphicx}
\usepackage{amsfonts}
\usepackage{amsmath}
\usepackage{color}
\usepackage{booktabs}
\usepackage{multirow}
\usepackage{array}
\usepackage{colortbl}
\usepackage{xcolor}
\usepackage{caption}

\usepackage{bbding}

\title{Improved Masked Image Generation with Knowledge-Augmented Token Representations}
\author{
    Guotao Liang\textsuperscript{\rm 1}\textsuperscript{\rm 2}, Baoquan Zhang\textsuperscript{\rm 1}\thanks{Corresponding Authors.}, Zhiyuan Wen\textsuperscript{\rm 2}, Zihao Han\textsuperscript{\rm 1}, Yunming Ye\textsuperscript{\rm 1}\textsuperscript{\rm 2}
}
\affiliations{
    \textsuperscript{\rm 1}Harbin Institute of Technology, Shenzhen;
    \textsuperscript{\rm 2}Peng Cheng Laboratory;\\

    lianggt@pcl.ac.cn, \{baoquanzhang, yeyunming\}@hit.edu.cn, zihaohan@stu.hit.edu.cn, \\
    wenzhiyuan2012@gmail.com
}

\usepackage{bibentry}

\begin{document}

\maketitle

\begin{abstract}
Masked image generation (MIG) has demonstrated remarkable efficiency and high-fidelity images by enabling parallel token prediction. Existing methods typically rely solely on the model itself to learn semantic dependencies among visual token sequences. However, directly learning such semantic dependencies from data is challenging because the individual tokens lack clear semantic meanings, and these sequences are usually long. To address this limitation, we propose a novel Knowledge-Augmented Masked Image Generation framework, named KA-MIG, which introduces explicit knowledge of token-level semantic dependencies (\emph{i.e.}, extracted from the training data) as priors to learn richer representations for improving performance. In particular, we explore and identify three types of advantageous token knowledge graphs, including two positive and one negative graphs (\emph{i.e.}, the co-occurrence graph, the semantic similarity graph, and the position-token incompatibility graph). Based on three prior knowledge graphs, we design a graph-aware encoder to learn token and position-aware representations. After that, a lightweight fusion mechanism is introduced to integrate these enriched representations into the existing MIG methods. Resorting to such prior knowledge, our method effectively enhances the model's ability to capture semantic dependencies, leading to improved generation quality. Experimental results demonstrate that our method improves upon existing MIG for class-conditional image generation on ImageNet. 
\end{abstract}

\begin{links}
    \link{Code}{https://github.com/GuotaoLiang/KA-MIG}
\end{links}

\section{Introduction}
Class-conditional images are a fundamental task in generative modeling, which aims to synthesize realistic images conditioned on given semantic class labels. Traditional methods are mainly based on generative adversarial networks (GANs) \cite{brock2018large,sauer2022stylegan,kang2023scaling} 
which have achieved impressive visual quality and class-conditional controllability. In recent years, inspired by next-token prediction in natural language processing \cite{vaswani2017attention}, the autoregressive generation paradigm based on discrete tokens has emerged as a promising alternative \cite {razavi2019generating,esser2021taming,sun2024autoregressive}. However, these methods often suffer from slow sampling and degraded image quality due to the long visual token sequences.
\begin{figure}[t]
    \centering
    \includegraphics[width=1\linewidth]{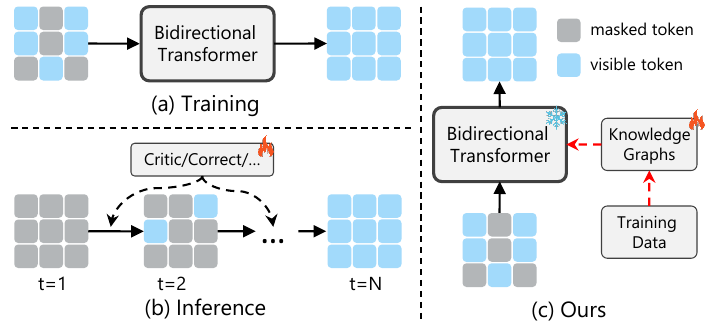}
    \caption{Motivation and framework overview. (a) The training pipeline of Masked Image Generation (MIG). (b) The inference process in MIG, where existing works primarily focus on improving sampling quality. (c) Our proposed framework incorporates external prior knowledge graphs to learn richer representations for enhancing MIG.}
    \label{fig:motivation}
    \vspace{-2mm}
\end{figure}

Recent advances in masked image generation (MIG) frameworks, such as MaskGIT~\cite{chang2022maskgit}, formulate image synthesis as a parallel decoding process that iteratively predicts masked tokens. This approach achieves a favorable trade-off between sampling speed and quality, and has become a representative paradigm in token-based image generation. However, it still underperforms recent well-developed diffusion-based models~\cite{rombach2022high}. To address this, many novel methods have been proposed to improve the generation performance, including introducing criticism or correction strategies~\cite{lezama2022improved,lezama2022discrete}, or proposing self-guidance sampling method~\cite{hur2024unlocking}, or devising a more uniform sampling strategy~\cite{besnier2025halton}. While these approaches have made notable progress, they primarily focus on refining the decoding strategy, leaving the intrinsic representation and modeling capacity of MaskGIT itself largely underexplored.

Existing methods mainly rely on transformer itself to learn visual token dependencies. However, directly learning such semantic dependencies from training data is challenging for two main reasons. First, individual tokens lack clear semantic meaning, making it difficult for the model to interpret or relate them effectively. Second, visual token sequences are typically long (\emph{e.g.}, 256 tokens per image), which hinders the capture of complex relations and leads to training inefficiencies and unstable generation performance.

Motivated by the above limitation, we propose a novel \textit{\textbf{K}nowledge-\textbf{A}ugmented} Masked Image Generation framework, called KA-MIG. The novelty lies in leveraging prior semantic knowledge graphs (\emph{i.e.}, token-level relations) to guide the learning of richer token representations for improving generation performance. However, as mentioned above, each token lacks interpretable semantics in the dataset, making it difficult to obtain such knowledge graphs. To handle this, we adopt a data-driven approach to uncover latent structural regularities from large-scale training data. Its advantage lies in its efficiency and compatibility, as it requires neither external annotations nor manually designed rules. Follow this, we construct three types of advantageous token knowledge graphs, consisting of two positive priors and one negative prior: 1) co-occurrence graph, which captures frequent co-appearances of token pairs and reflects latent spatial-semantic correlations; 2) semantic similarity graph, which identifies tokens with analogous meanings in the context of image synthesis; 3) position-token incompatibility graph, which captures negative prior, identifying tokens that are incompatible with certain spatial locations conditioned on class labels. The incompatible means that tokens that have never appeared at that position. For example, in the ``airplane'' class, tokens representing ground or grass textures seldom occur in the upper-middle image region, as that space is predominantly occupied by the airplane body or sky. Building on these graphs, a lightweight fusion framework is introduced to integrate the enriched prior knowledge into the existing MIG methods. More specifically, for positive priors, we inject the prior features into each layer of the transformer in an additive manner. Different from positive priors, negative prior knowledge is incorporated using a subtractive manner. This simple yet effective fusion scheme guides the model towards capturing richer semantic dependencies and spatial constraints without introducing significant computational complexity.

The contributions of this work are summarized as follows:
\begin{itemize}
    \item Unlike existing works that focus on improving sampling strategies, we propose a novel knowledge-augmented masked image generation framework, named KA-MIG, which leverages prior semantic knowledge graphs to enrich internal token representations and improve generation performance.
    \item We systematically construct three types of advantageous token knowledge graphs, consisting of two positive priors and one negative prior, \emph{i.e.}, co-occurrence graph, semantic similarity graph, and position-token incompatibility graph. Resorting to such prior knowledge, we introduce a lightweight additive–subtractive fusion mechanism to effectively integrate both positive and negative knowledge into the existing framework.
    \item Extensive experiments show that KA-MIG significantly enhances the generation performance of existing MIG models, demonstrating its effectiveness and versatility.
\end{itemize}

\begin{figure*}[th]
    \centering
    \includegraphics[width=1\linewidth]{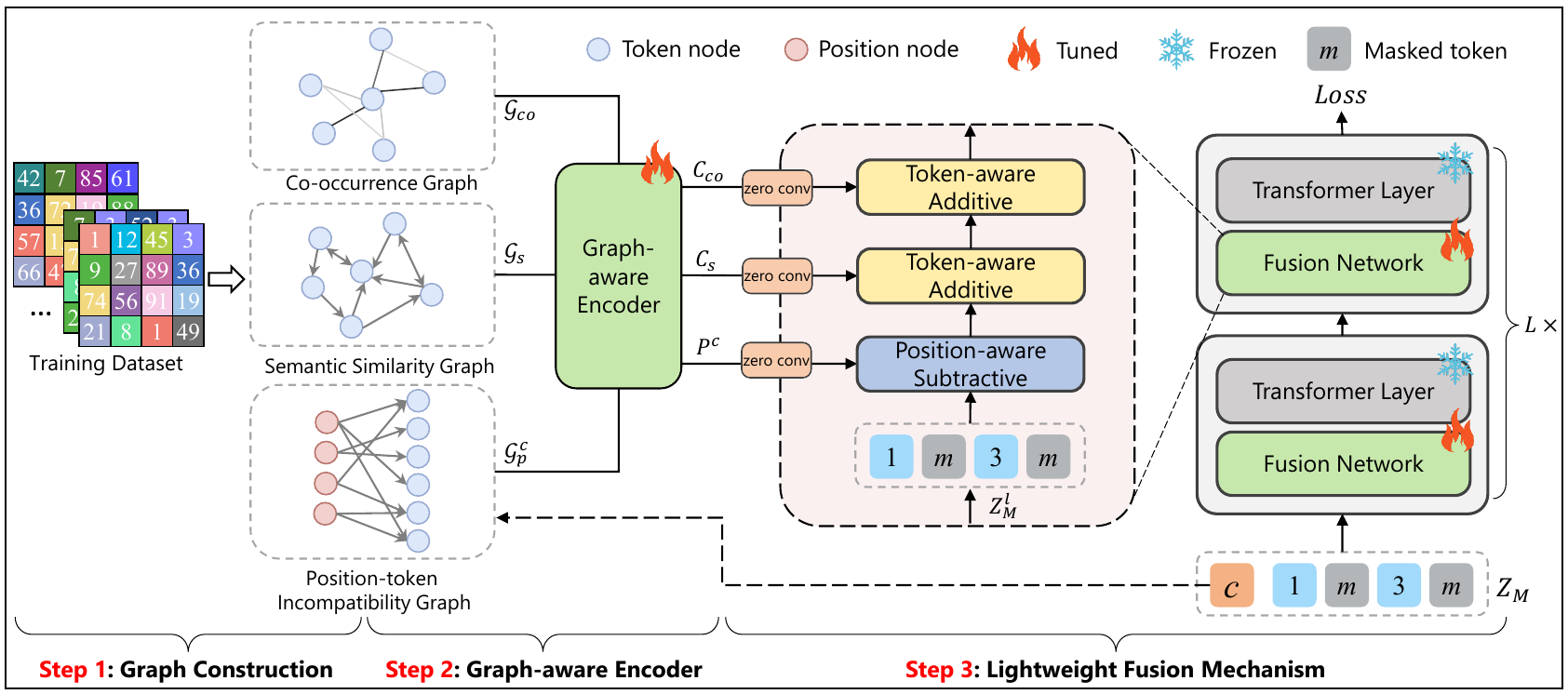}
    \caption{The illustration of our proposed KA-MIG framework. We first construct three types of prior knowledge graphs from the training dataset, \emph{i.e.}, $\mathcal{G}_{co}$, $\mathcal{G}_{s}$, and $\mathcal{G}_{p}^{c}$. These graphs are then fed into a graph-aware encoder to learn richer token- and position-aware representations, \emph{i.e.}, $C_{co}$, $C_{s}$, and $P^{c}$. Finally, the enriched prior representations are integrated into the existing MIG framework through a lightweight fusion network.}
    \label{fig:model}
\end{figure*}
\section{Related Works}
\textbf{Image generation}, the task of synthesizing realistic images from noise or structured inputs, has long been a core challenge in computer vision. Generative adversarial networks (GANs) initially dominated image generation research~\cite{goodfellow2014generative,brock2018large}. Despite their success, GANs suffer from training instability, mode collapse, and difficulty in modeling diverse data distributions~\cite{sauer2022stylegan,kang2023scaling,miyato2018spectral}. Motivated by this limitation, \textit{diffusion-based models}~\cite{dhariwal2021diffusion,rombach2022high,han2025asyncdsb} and \textit{token-based models}~\cite{esser2021taming} have recently emerged as powerful alternatives. \textit{Diffusion-based models} formulate image generation as progressive denoising process, starting from pure Gaussian noise and learning to reconstruct images through a series of iterative refinements.

Instead of directly generating pixel values, \textit{token-based models} first rely on vector quantization techniques~\cite{van2017neural} to compress continuous images into sequences of discrete tokens, and then learn to model and generate these token sequences. Two primary prediction paradigms exist: autoregressive and non-autoregressive approaches. Autoregressive models generate tokens sequentially, \emph{i.e.}, predicting next token based on previously generated tokens~\cite{esser2021taming,guotao2024lg,liang2025towards}. 

Non-autoregressive models, \emph{i.e.}, masked image generation, generate multiple tokens per step, enabling parallel decoding. A representative work is MaskGIT~\cite{chang2022maskgit}, which predicts the entire token map in parallel and iteratively refines it using confidence-based masking. This results in significantly faster inference with competitive generation quality. 
However, it still underperforms recent well-developed diffusion-based models. Unfortunately, efforts to improve its performance remain relatively limited. Existing works focus on improving the inference sampling strategy. For example, training a critic or correctors post-hoc to correct tokens with intermediate decoding errors~\cite{lezama2022improved,lezama2022discrete}. Some works~\cite{ni2024revisiting,ni2024adanat} attempt to directly optimize training and sampling hyper-parameters to obtain better performance. Inspired by the self-guidance approach in diffusion models, SG-MGM~\cite{hur2024unlocking} adapts the self-guidance approach to Masked Image Generation (MIG) to address the multi-modality problem~\cite{gu2017non}. A recent work~\cite{besnier2025halton} argues that the confidence scheduling strategy of the MaskGIT method impacts the generated images’ diversity and quality, and proposes a new sampling strategy based on Halton low-discrepancy sequence~\cite{halton1964radical}, which can reveal new tokens more uniformly.

Unlike prior methods that mainly refine the decoding strategy, our approach focuses on improving model's internal representation capacity by modeling richer semantic dependencies among tokens. We identify three types of advantageous token knowledge graphs and propose a lightweight additive–subtractive fusion mechanism to integrate prior knowledge into the existing framework effectively.

\section{Preliminaries: MIG}
Mask image generation (MIG) relies on a pre-trained VQ-VAE model~\cite{van2017neural} that compresses continuous images into a discrete token sequence. The VQ-VAE consists of an encoder $E_{vq}$, a quantizer $Q$ with a learnable codebook $\mathcal{C}$, and a decoder $D_{vq}$. Given an image $x \in \mathbb{R}^{H \times W \times 3}$, the encoder and quantizer compress the image into a discrete token sequence: $Z = Q(E_{vq}(x)), Z = [z_i]_{i=1}^{N}$, where $z_i$ is a specific entry in codebook, $N$ is the length of the sequence. The decoder is used to reconstruct the original image based on $Z$. MIG aims to learn to generate reasonable discrete token sequences. \\
\textbf{Training strategy}. Give an external condition $c$,  MIG first samples a binary mask matrix $M = [m_i]_{i=1}^{N}$. This token $z_i$ is replaced by the special token $[\text{MASK}]$ if $m_i=1$, otherwise $z_i$ will keep unchanged when $m_i=0$. MIG, a multi-layer bidirectional transformer $p_{\theta}(\cdot)$, is trained to restore the masked tokens to clean tokens based on unmasked tokens. The training objective is to minimize the negative log-likelihood of the masked tokens:
\begin{equation}
    \mathcal{L_{\text{MIG}}} = - \mathop{\mathbb{E}}\limits_{ Z \in \mathcal{D}} \left[ \sum_{\forall i \in [1, N],\, m_i = 1} \log p_{\theta}(z_i \mid Z_{\overline{M}}, c) \right].
\label{eq:mig_loss}
\end{equation}\\
\textbf{Generation strategy}. MIG starts from a fully masked token sequence and iteratively predicts all masked positions in parallel. At each step, a subset of tokens with the highest confidence scores are selected and kept, gradually completing the sequence over multiple iterations.

\section{Proposed Method: KA-MIG}
Existing works attempt to refine sampling strategies to improve generation performance, while ignoring a fundamental limitation that the model itself struggles to effectively capture semantic dependencies among visual tokens because individual tokens lack clear semantic meanings. To address this, we propose KA-MIG, a novel framework that introduces explicit knowledge of token-level semantic dependencies (\emph{i.e.}, extracted from the training data) as priors to guide the learning of richer token representations. As illustrated in Figure~\ref{fig:model}, our KA-MIG consists of three key steps, \emph{i.e.}, graph construction, graph-aware encoder, and lightweight fusion mechanism. Here, we elaborate on each step in order:

\textbf{Step 1: \emph{Graph Construction}}. Due to the absence of explicit semantic meanings assigned to individual tokens in the dataset, this poses a challenge for us to construct prior knowledge graphs. To address this, we seek to uncover latent structural regularities in a data-driven manner. Specifically, we construct three types of advantageous graphs, consisting of two positive priors and one negative prior:

\textit{Co-occurrence Graph $\mathcal{G}_{co}$} indicates that two tokens are likely to co-occur in a local range, which reflects the latent spatial-semantic correlations. Such correlations have been shown to be effective in graph learning~\cite{nguyen2022node} and recommendation systems~\cite{han2022multi}. Specifically, we construct a weighted, undirected token-to-token graph $\mathcal{G}_{co}$. Each edge $\mathcal{E}_{ij}^{co}$ between token $i$ and token $j$ records how often the two tokens co-occur in a first-order local neighborhood (\emph{i.e}, vertical, horizontal, and diagonal directions) within the training data. To reduce noise, we prune the graph by removing low-frequency edges.

\begin{figure}[t]
    \centering
    \includegraphics[width=1\linewidth]{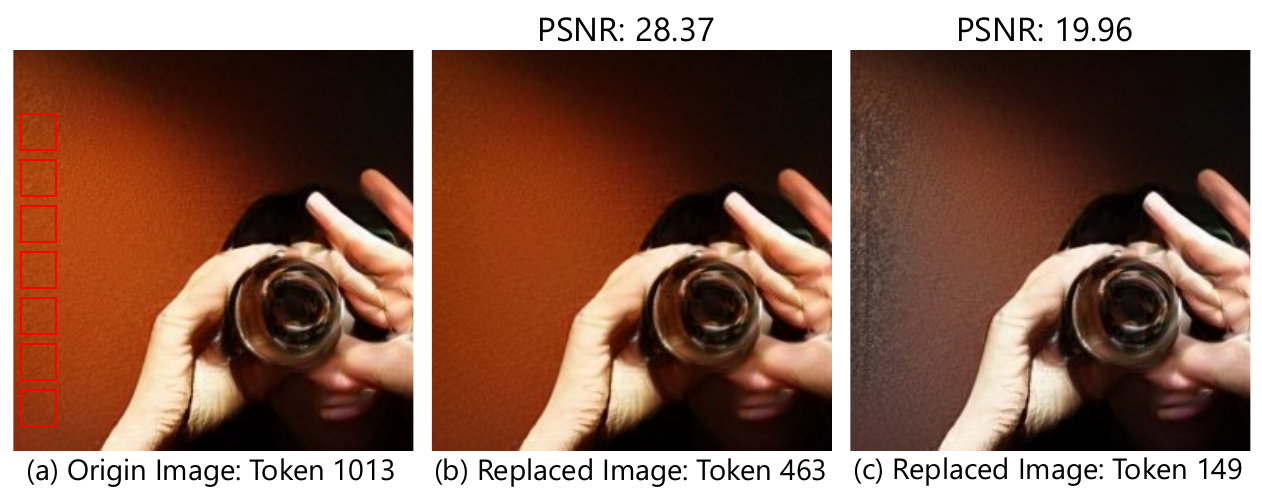}
    \caption{Visualization of semantic similarity token via reconstruction. Peak Signal-to-noise Ratio (PSNR)~\cite{fardo2016formal} measures the pixel-level similarity between two images. The red box indicates the location of token (1013). We compare three versions of the same image: (a) the original reconstruction, (b) replacing token (1013) with its most similar token (463) based on our similarity graph $\mathcal{G}_s$, and (c) replacing it with the least similar token (149). The perceptual fidelity of (b) supports the semantic closeness, while (c) introduces noticeable degradation. }
    \label{fig:sim_graph_example}
    \vspace{-5mm}
\end{figure}

\textit{Semantic Similarity Graph $\mathcal{G}_{s}$} indicates that two tokens express similar semantics in the context of image synthesis, which is akin to the definition of synonyms in natural language. Its advantage is that it encourages the model to capture interchangeable token patterns and improve robustness. However, constructing this graph is very challenging due to the absence of explicit semantic meanings for individual tokens. 
To capture such patterns, we hypothesize that tokens with similar positional distributions across a wide range of images are likely to convey similar semantics. Based on this, for each token, we construct a position distribution vector of length $N$, where each entry represents the frequency of the token appearing at a specific spatial position in the training images. We then compute the Jensen-Shannon (JS) divergence between position distribution vectors to quantify token similarity. 
For each token, we retain the top two most similar tokens to form a directed token-to-token graph $\mathcal{G}_{s}$.
To validate our approach, as illustrated in Figure~\ref{fig:sim_graph_example}, we replace token (1013) with its most similar token (463) and its least similar token (149), and input the modified token maps into the decoder for image reconstruction. The results show that replacing with token (463) yields no perceptible difference from the original image, whereas token (149) significantly degrades image quality. 

\textit{Position-token Incompatibility Graph $\mathcal{G}_{p}$} serves as a negative prior, by identifying tokens that are incompatible with specific spatial locations under particular class labels. For example, in the “airplane” class, tokens representing ground or grass textures rarely appear in the upper-middle region of the image, as the airplane body or sky typically occupies that area. Incorporating such a negative prior helps the model avoid implausible compositions, resulting in more semantically consistent and spatially coherent image synthesis. Specifically, for each class $c$, we construct a directed position-to-token graph $\mathcal{G}_{p}^{c}$, where each entry $\mathcal{E}_{ij}^{pc} = 1$ denotes that token $j$ is incompatible with spatial position $i$. To construct this graph, we scan all training images of class $c$ and record tokens that never appear at each spatial location.

\textbf{Step 2: \emph{Graph-aware Encoder}} After obtaining the prior graphs described above, our goal is to learn richer representations from them for injection into the generation process, thereby enhancing image quality. To achieve this, we design two distinct encoders to process positive and negative priors. 
For the positive priors $\mathcal{G}_{co}$ and $\mathcal{G}_{s}$, the goal is to learn global token representations that reflect semantic dependencies among tokens. To this end, we employ two graph convolutional networks $f_{\theta_{co}}$ and $f_{\theta_{s}}$, to extract global token representations, respectively:
\begin{equation}
\begin{aligned}
     C_{co} &= f_{\theta_{co}}(\mathcal{G}_{co}, C), \\
     C_{s} &= f_{\theta_{s}}(\mathcal{G}_{s}, C),
\end{aligned}
\end{equation}
where $C$ is the codebook embedding. 

For the negative prior, the objective is to learn position-specific features that reflect which tokens are unlikely to occur at specific position. Specifically, given class $c$, for each position $p_{i}^{c}$, we collect the set of incompatible tokens $\mathcal{I}_{i,j}$ from $\mathcal{G}_{p}^{c}$, and compute its representation by averaging the corresponding token embeddings:
\begin{equation}
    p_{i}^{c} = \frac{1}{|\mathcal{I}_{i,j}|} \sum_{t \in \mathcal{I}_{i,j}} C_t W,
\end{equation}
where $C_t$ is the embedding of token $t$ in the codebook $C$, $W$ is a learnable weight matrix. This yields a position embedding $P^{c} \in \mathbb{R}^{N \times d}$ that encodes negative spatial constraints.\\
\begin{figure*}
    \centering
    \includegraphics[width=1\linewidth]{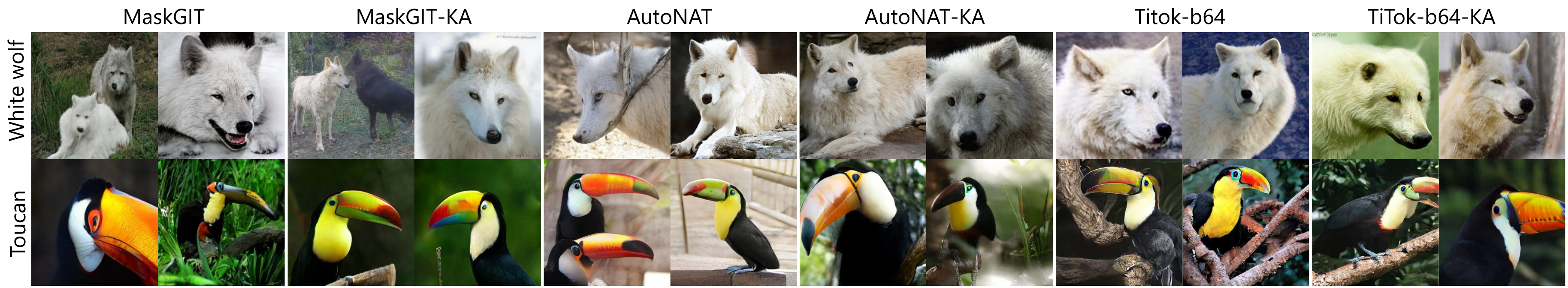}
    \caption{Visualizations for sampling images on ImageNet-256 using selected classes (270: White wolf, and 96: Toucan). }
    \label{fig:visual_img}
    \vspace{-3mm}
\end{figure*}
\textbf{Step 3: \emph{Lightweight Fusion Mechanism}} Considering that existing MIG models have already captured rich knowledge from large-scale training, it is crucial to carefully integrate additional prior information so as not to interfere with the learned representations. Motivated by this, a lightweight fusion mechanism is proposed to solve this, as shown in Figure~\ref{fig:model}. Specifically, given class $c$, input masked sequence $Z_{M}$, and a pre-trained MIG, a fusion network is inserted before the input of each transformer layer of MIG. For the positive priors $C_{co}$ and $C_{s}$, we enhance the unmasked token representations $Z_{\overline{M}}$ by additive operations at $l$-th layer:
\begin{equation}
    Z_{\overline{M}}^{l} = Z_{\overline{M}}^{l} + f_{pos}^{l}(C_{co} [Z_{\overline{M}}]) + f_{pos}^{l} (C_{s} [Z_{\overline{M}}]),
\end{equation}
where $C_{co} [Z_{\overline{M}}]$ and $C_{s} [Z_{\overline{M}}]$ denote the unmask token embedding retrieved from the global prior codebook embeddings, $f_{pos}(\cdot)$ is a zero convolution mapping function~\cite{zhang2023adding}.

In contrast, for the negative prior $P^{c}$, we suppress the token features at each position by subtraction:
\begin{equation}
    Z_{M}^{l} = Z_{M}^{l} - \alpha f_{neg}^{l} (P^{c}),
\end{equation}
where $f_{neg}(\cdot)$ is also implemented as a zero convolution function, $\alpha$ is a hyperparameter.

\begin{table*}[th]
\centering

\begin{tabular}{lccccccc}
\bottomrule
\multicolumn{1}{c}{\multirow{2}{*}{Model}}  & \multirow{2}{*}{Type} & \multicolumn{6}{c}{ImageNet 256$\times$256}                                          \\ \cline{3-8} 
\multicolumn{1}{c}{}                        &                       & \#Params & Steps & FID$\downarrow$ & IS$\uparrow$   & Prec$\uparrow$ & Rec$\uparrow$ \\ \hline
BigGAN-deep (Brock et al. 2018)             & GANs                  & 112M     & 1     & 6.95            & 224.5          & \textbf{0.89}  & 0.38          \\
GigaGAN~\cite{kang2023scaling}              & GANs                  & 569M     & 1     & 3.45            & 225.5          & 0.84           & 0.61          \\ \hline
ADM~\cite{dhariwal2021diffusion}            & Diff.                 & 554M     & 250   & 10.94           & 101.0          & 0.69           & \textbf{0.63} \\
CDM~\cite{ho2022cascaded}                   & Diff.                 & $-$      & 8100  & 4.88            & 158.7          & $-$            & $-$           \\
LDM-4~\cite{rombach2022high}              & Diff.                 & 400M     & 250   & 3.60            & 247.7          & $-$            & $-$           \\ \hline
VQVAE-2 (Razavi et al., 2019)               & AR                    & 13.5B    & 5120  & 31.11           & 45             & 0.36           & 0.57          \\
VQGAN (Esser et al. 2021)                   & AR                    & 227M     & 256   & 18.65           & 80.4           & 0.78           & 0.26          \\
LlamaGen-Xl~\cite{sun2024autoregressive} & AR                    & 775M     & 576   & 2.62            & 244.1          & 0.80           & 0.57          \\
LlamaGen-XXl~\cite{sun2024autoregressive} & AR                    & 1.4B     & 576   & 2.34            & 253.9          & 0.80           & 0.59          \\
VAR-d20~\cite{tian2024visual}               & AR                    & 600M     & 10    & 2.57            & 302.6          & 0.83           & 0.56          \\
VAR-d30~\cite{tian2024visual}               & AR                    & 2.0B     & 10    & 1.92            & \textbf{323.1}          & 0.82           & 0.59          \\ \hline
Token-Critic~\cite{lezama2022improved}      & MIG                 & 422M     & 36    & 4.69            & 174.5          & 0.76           & 0.53          \\
DPC-light~\cite{lezama2022discrete}         & MIG                 & $-$      & 66    & 4.8             & 249.0          & 0.80           & 0.50          \\
DPC-full~\cite{lezama2022discrete}          & MIG                 & $-$      & 180   & 4.45            & 244.8          & 0.78           & 0.52          \\
MaskGIT-SAG~\cite{hur2024unlocking}         & MIG                 & $-$      & 12    & 3.35            & 259.7          & 0.81           & 0.52          \\
MaskGIT-Halton~\cite{besnier2025halton}     & MIG                 & 705M     & 32    & 3.74            & 279.5          & 0.81           & 0.60          \\ \hline
MaskGIT~\cite{chang2022maskgit}             & MIG                 & 227M     & 8     & 6.18            & 182.1          & 0.80           & 0.52          \\ \rowcolor{gray!30}
MaskGIT-KA (Ours)                            & MIG                 & 245M     & 8     & 5.69   & 170.2          & 0.81           & 0.50          \\  
AutoNAT~\cite{ni2024revisiting}             & MIG                 & 194M     & 8     & 2.68            & 278.8          & $-$            & $-$           \\ \rowcolor{gray!30}
AutoNAT-KA (Ours)                            & MIG                 & 211M     & 8     & 2.45   & 274.1          & 0.82           & 0.56          \\
TiTok-b64~\cite{yu2024image}                & MIG                 & 177M     & 8     & 2.48            & 214.7          & $-$            & $-$           \\ \rowcolor{gray!30}
TiTok-b64-KA (Ours)                             & MIG                 & 194M     & 8     & 2.40   & 217.0 & 0.78           & 0.60          \\
TiTok-s128~\cite{yu2024image}               & MIG                 & 177M     & 64     & 1.97            & 281.8          & $-$            & $-$           \\ \rowcolor{gray!30}
TiTok-s128-KA (Ours)                            & MIG                 & 194M     & 64     & \textbf{1.90}            & 271.9          & 0.78           & 0.61 \\ \toprule        
\end{tabular}
\caption{Class-conditional image generation on ImageNet-256. Our method is marked with \colorbox{gray!30}{gray}.  We calculate FID-50K following~\cite{dhariwal2021diffusion}. The best results are highlighted in \textbf{bold}. Diff: diffusion, AR: autoregressive.}
\label{tab:generation_exp}
\vspace{-2mm}
\end{table*}

\subsection{Training and Generation} \label{sec:train_generation}
Following the above process, we adopt Equation~\ref{eq:mig_loss} as the training objective to optimize our model. Since the prior knowledge is learned independently of the MIG framework, the corresponding representations can be precomputed and stored. During inference, only lightweight additive and subtractive operations are performed, thereby incurring minimal computational overhead. More detailed analysis can be found in Section Ablation Study. This makes our method both practical for real-world deployment and fully compatible with existing MIG frameworks.

\section{Experiments}
\subsection{Experimental Settings}
\textbf{Benchmarks and Datasets}. As our method is model-agnostic, allowing it to be applied to different MIG architectures. Then, we choose MaskGIT \cite{chang2022maskgit}, AutoNAT \cite{ni2024revisiting}, and recently state-of-the-art work TiTok~\cite{yu2024image} as our backbone networks. We evaluate the effectiveness of our method in class conditional generation using the ImageNet benchmark~\cite{deng2009imagenet}. Due to AutoNAT and TiTok not having publicly available checkpoints on 512$\times$512, we only evaluated them at 256$\times$256.  \\
\textbf{Evaluation Metrics}. Following previous works \cite{dhariwal2021diffusion,chang2022maskgit, ni2024revisiting, yu2024image, hur2024unlocking}, we adopt the standard Fréchet Inception Distance (FID) \cite{heusel2017fid}, Inception Score (IS)~\cite{salimans2016improved}, Precision, and Recall~\cite{kynkaanniemi2019improved} to evaluate the trade-off between sample fidelity and diversity. \\
\textbf{Implementation Details}. Our training and generation configurations follow the backbone model with minor adjustments to batch sizes, learning rates, and weight regularization to obtain better performance. All knowledge graphs are normalized. The $f_{\theta_{co}}$ and $f_{\theta_{s}}$ are 3-layer GCNs with the ReLU activation function. We freeze the backbone network and only fine-tune the classification layer and the additional parameters introduced by our method. More detailed experimental settings can be found in the supplementary material.
\begin{table}[t]
\footnotesize
\centering
\begin{tabular}{lccccc}
\bottomrule
        & Step & FID$\downarrow$ & CLIP-Score $\uparrow$   \\ \hline
MaskGen~\cite{kim2025democratizing} & 16   & 22.27 & 25.58 \\
Ours    & 16   &\textbf{21.01}	&\textbf{26.10}      \\ \toprule
\end{tabular}
\caption{Text-to-image generation on COCO-30K.}
\label{tab:t2i}
\vspace{-4mm}
\end{table}
\subsection{Discussion of Results}
\textbf{ImageNet-256} In Table~\ref{tab:generation_exp}, we present a comparison of our method against the state-of-the-art generative models and the backbone models on ImageNet-256. The comparison models involve GAN-based, diffusion-based models, recently proposed state-of-the-art autoregressive models, and models with improved MIG sampling strategies. From the results, we can obtain several key conclusions:
1) Compared with the backbone models, our method consistently enhances generation performance, demonstrating both its effectiveness and versatility. This improvement is attributed to the enriched internal representations brought by incorporating prior semantic knowledge graphs, which help the model better capture inter-token dependencies; 
2) Compared with MaskGIT-KA and AutoNAT-KA, the performance improvement of TiTok-KA is relatively marginal. This result is expected and reasonable. On the one hand, TiTok uses a significantly shorter token sequence (64 or 128 tokens) than MaskGIT and AutoNAT (256 tokens). Longer sequences entail more complex token interactions. Then, integrating such prior knowledge is helpful to improve the modeling ability in this aspect. This explains the more substantial improvements observed on MaskGIT and AutoNAT, supporting the validity of our core motivation. On the other hand, despite TiTok’s compact design, our method still yields consistent performance gains, further demonstrating the generalizability and robustness of the proposed approach.
3) Compared to GAN, Diff, and AR generative models, our TiTok-s128-KA achieves the best FID score, demonstrating the superiority of the proposed approach. Moreover, the AutoNAT model achieves more competitive performance after integrating our method. For example, in FID, AutoNAT-KA achieves a score of 2.45, outperforming LlamaGen-XL (2.62), VAR-d20 (2.57), and TiTok-b64 (2.48). These results further validate the effectiveness and general applicability of our method; 
4) Our method introduces only approximately 20 million additional parameters, demonstrating its lightweight design and efficiency.
Finally, we also provide a qualitative comparison of image generation in Figure~\ref{fig:visual_img}. \\
\textbf{Text-to-image generation on MS-COCO.} We further evaluate the effectiveness of our method in the text-to-image generation scenario, using the MS-COCO~\cite{lin2014coco} benchmark. We select MaskGen~\cite{kim2025democratizing} as our backbone. Since the $\mathcal{G}_{p}$ is customized to the class condition, we only incorporate $\mathcal{G}_{co}$ and $\mathcal{G}_{s}$ into MaskGen. The results are summarized in Table~\ref{tab:t2i}, which demonstrates the effectiveness and versatility of our method. \\

\begin{figure*}[h]
    \centering
    \includegraphics[width=1\linewidth]{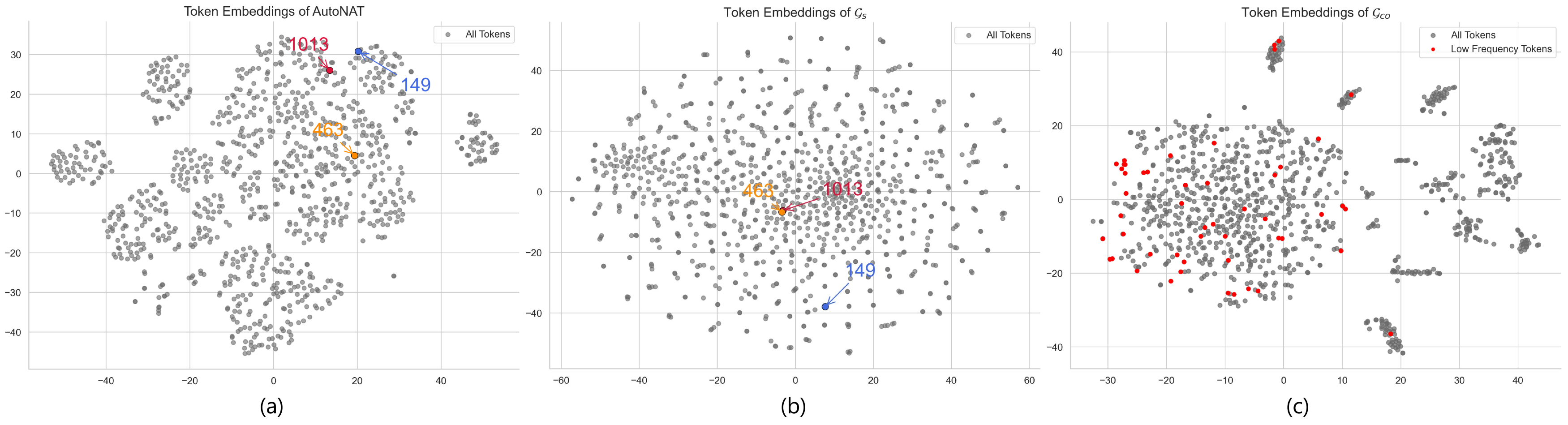}
    \caption{Visualization of token embeddings from (a) AutoNAT, (b) semantic similarity graph, and (c) co-occurrence graph.}
    \label{fig:token_emb}
    \vspace{-2mm}
\end{figure*}
\begin{table}[t]
\centering
\begin{tabular}{llcc}
\bottomrule
      &                                                              & FID$\downarrow$ & IS$\uparrow$ \\ \hline
(i)   & Baseline (AutoNAT)                                            & 2.68            & 278.8       \\
(ii) & + $\mathcal{G}_{s}$                                          & 2.49            & 279.6            \\
(iii)  & + $\mathcal{G}_{p}$                                          & 2.51            & 285.6    \\
(iv)  & + $\mathcal{G}_{co}$                                         & 2.51            & 282.1            \\
(v)  & + $\mathcal{G}_{s}$  +  $\mathcal{G}_{p}$                     & 2.46               &  279.9           \\
(vi)   & + $\mathcal{G}_{co}$ + $\mathcal{G}_{p}$                     & 2.46               &  280.7             \\
(vii)  & + $\mathcal{G}_{co}$ + $\mathcal{G}_{s}$                     & 2.48               &  277.4       \\ 
(viii)  & + $\mathcal{G}_{co}$ + $\mathcal{G}_{s}$ + $\mathcal{G}_{p}$ & 2.45            & 274.1       \\ \toprule
\end{tabular}
\caption{Ablation study of three graphs on ImageNet-256. ``$+ \mathcal{G}_{co}$'' denotes introducing co-occurrence graph.}
\label{tab:ablation_study}
\vspace{-5mm}
\end{table}
\begin{figure}[t]
    \centering
    \includegraphics[width=1\linewidth]{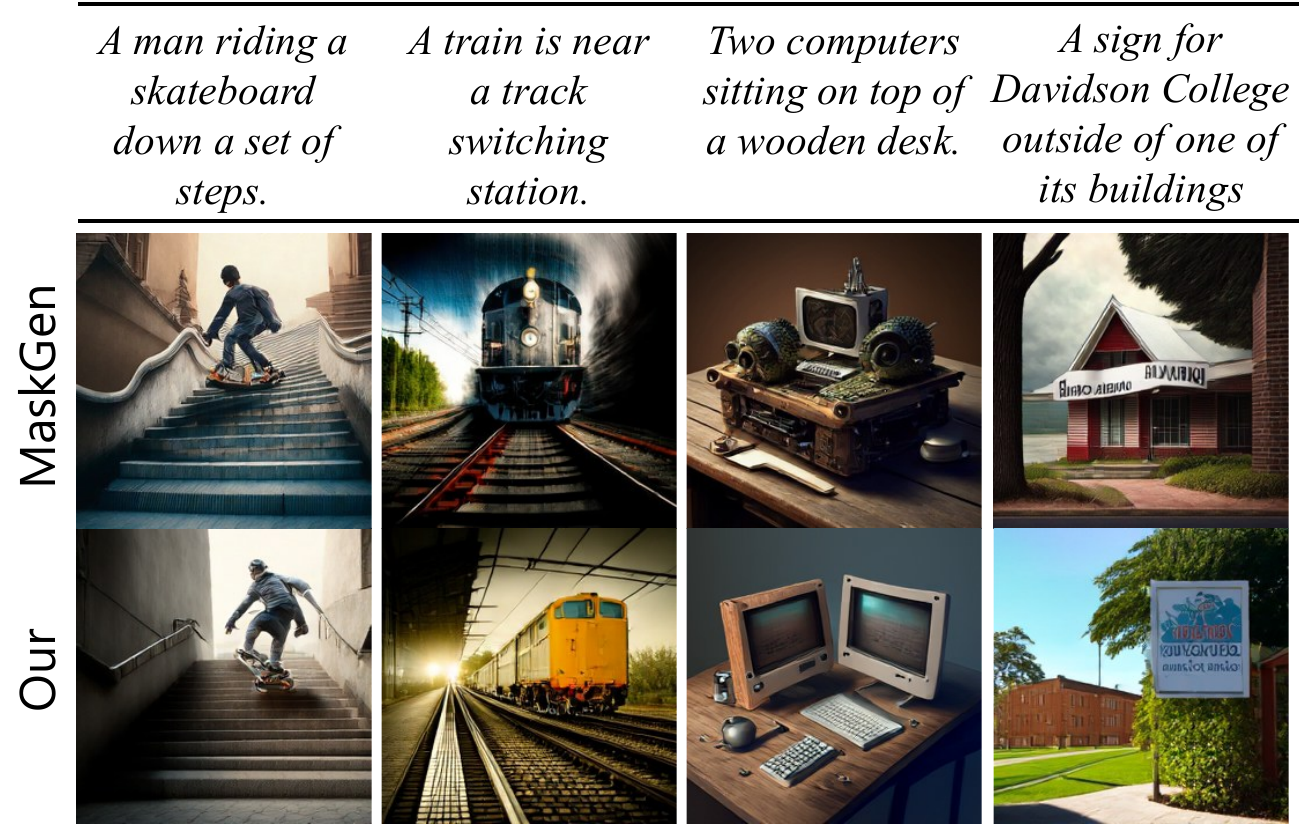}
    \caption{Visualizations for text-to-image.}
    \label{fig:text2image}
    \vspace{-5mm}
\end{figure}
\subsection{Ablation Study}
\textbf{Are all three prior knowledge graphs both effective?} To investigate this, we conduct extensive ablation studies to evaluate the individual and combined impact of the three graphs. We use AutoNAT as the backbone model, and the results are presented in Table~\ref{tab:ablation_study}. The results lead to several key observations:
First, each graph plays a crucial role in improving performance, which is reasonable because each captures semantic dependencies among tokens from different perspectives, thereby validating the rationale behind our design.
Second, by comparing (ii) to (iv), we observe that $\mathcal{G}_{s}$ leads to the most significant performance improvement. This can be attributed to its ability to guide the model toward learning interchangeable token patterns, which helps enhance the robustness and diversity of the generated results.
Third, compared to the use of individual graphs in (ii) to (iv), combining any two graphs (\emph{i.e.}, (v) to (vii)) leads to further performance gains. This indicates that the prior graphs are highly complementary, contributing to richer and more diverse representations.
Finally, comparing the results of (viii) with those of (i) to (vii) shows that incorporating all three graphs yields the best overall performance, further demonstrating the effectiveness of our proposed approach. \\
\textbf{Efficiency Analysis.} As described above, our method is independent of the MIG framework, which means that graph features can be precomputed to reduce computational overhead. To demonstrate this, we conduct a detailed analysis in Table~\ref{tab:efficiency} to evaluate the impact of different storage strategies on model parameters and TFLOPs for the three graphs. We compute the TFLOPs of the entire generation process (excluding the VQ decoder) using an NVIDIA RTX 4090 GPU. 
``Computed Online'' denotes that the prior graph features are computed dynamically during inference, whereas ``Precomputed'' indicates that the features are calculated offline prior to inference and cached for reuse. Observed from the results, compared to online computation, precomputing the $\mathcal{G}_{co}$ and $\mathcal{G}_{s}$ introduces minimal additional parameters (only +0.79M each) and virtually no extra TFLOPs, making them highly suitable for pre-inference caching. 
In contrast, $\mathcal{G}_{p}$ introduces substantial storage overhead (+196M) due to its class-conditional nature—each class must store its own graph. To mitigate this, we compute $\mathcal{G}_{p}$ online during inference, which incurs only a minor computational cost (+0.06 TFLOPs) while significantly reducing memory usage.
This hybrid configuration (marked with \checkmark~in the table) demonstrates that our approach is not only effective but also practical for real-world deployment, offering a favorable trade-off between performance and efficiency. \\
\textbf{How is the token embedding space learned from our semantic similarity graph?} To examine whether our method effectively captures semantic similarity relationships, we visualize the token embeddings learned from $\mathcal{G}_{s}$ using t-SNE, as shown in Figure~\ref{fig:token_emb} (b). 
The results show that many nodes form small clusters of two or three tokens, which aligns with our graph construction strategy, where each node connects to its top-2 most similar neighbors. More interesting, we observe that token (1013) is very similar to token (463) but is far from token (149), which is consistent with the example visualization in Figure~\ref{fig:sim_graph_example}. These observations demonstrate that our graph effectively captures fine-grained semantic relationships between tokens, further validating the rationale and effectiveness of introducing the $\mathcal{G}_{s}$. \\
\textbf{How is the token embedding space learned from our co-occurrence graph?} Following the above setting, we visualize the token embedding learned from $\mathcal{G}_{co}$ in Figure~\ref{fig:token_emb} (c). The visualization shows that tokens form multiple coherent clusters, indicating that tokens with strong co-occurrence relationships are mapped close to each other. This confirms that the graph effectively encodes meaningful semantic locality. Furthermore, low-frequency tokens (in red) are not isolated but embedded among semantically similar neighbors, demonstrating that rare tokens enable the learning of meaningful semantics through contextual information. \\
\textbf{Is $\mathcal{G}_p$ effective in reducing positional bias?} 
We conduct a token prediction experiment to demonstrate that $\mathcal{G}_p$ effectively improves the ranking of ground-truth tokens. We randomly sample 10,000 images, apply random masking, and input them into the model to predict the corresponding ground-truth tokens.  The results are presented in Table~\ref{tab:rank}, which shows that $\mathcal{G}_p$ improves the ranking of ground-truth tokens. This aligns with the design motivation of $\mathcal{G}_p$: by identifying and suppressing tokens that are incompatible with specific spatial positions under a given class label, the model is better guided toward ranking true tokens higher.

\begin{table}[t]
\centering
\footnotesize
\renewcommand\tabcolsep{2.25pt}
\begin{tabular}{lccc|ccc}
\bottomrule
Storage Type      & \multicolumn{3}{c|}{Computed Online}                           & \multicolumn{3}{c}{Precomputed}                             \\ \hline
Prior Graph       & $\mathcal{G}_{co}$ & $\mathcal{G}_{s}$ & $\mathcal{G}_{p}$     & $\mathcal{G}_{co}$ & $\mathcal{G}_{s}$ & $\mathcal{G}_{p}$  \\
Extra Params & +16M               & +16M              & +15M                  & +0.79M             & +0.79M            & +196M              \\
Extra TFLOPs      & +0.09              & +0.09             & +0.06                 & $\sim$0            & $\sim$0           & $\sim$0            \\
Selected          &        &       & \checkmark            & \checkmark         & \checkmark        &        \\ \toprule
\end{tabular}
\caption{Comparison of different storage and computation strategies for the three proposed prior graphs. Our selected configuration (marked with \checkmark) achieves a favorable trade-off by precomputing lightweight graphs ($\mathcal{G}_{co}$ and $\mathcal{G}_{s}$) to reduce runtime cost, while computing the heavier graph ($\mathcal{G}_{p}$) online to avoid excessive memory consumption. }
\label{tab:efficiency}
\vspace{-3mm}
\end{table}
\begin{table}[t]
\centering
\footnotesize
\begin{tabular}{lcc}
\bottomrule
                   & Mask Ratio & NDCG@100$\uparrow$ \\ \hline
AutoNAT            & 0.3        & 0.0282              \\
+$\mathcal{G}_{p}$ & 0.3        & \textbf{0.0284}     \\ \hline
AutoNAT            & 0.5        & 0.0288              \\
+$\mathcal{G}_{p}$ & 0.5        & \textbf{0.0289}     \\ \toprule
\end{tabular}
\caption{Comparison of NDCG@100 scores between the baseline model (AutoNAT) and our method with $\mathcal{G}_p$ under different mask ratios. The results show that $\mathcal{G}_p$ improves the ranking of ground-truth tokens.}
\label{tab:rank}
\vspace{-3mm}
\end{table}
\section{Conclusion}
In this work, we presented a novel Knowledge-Augmented Masked Image Generation framework that leverages explicitly constructed token-level semantic prior graphs to learn richer representations for improving performance. In particular, we uncover three types of advantageous token knowledge graphs, consisting of two positive priors and one negative prior: co-occurrence, semantic similarity, and position-token incompatibility graph. We also design a lightweight additive-subtractive fusion mechanism to integrate these priors into the existing framework effectively. Extensive experiments demonstrate the effectiveness and versatility of our approach across multiple backbone models.

\clearpage

\section*{Acknowledgement} 
This work was supported by the Guangdong Basic and Applied Basic Research Foundation under Grant No. 2025A1515011674, the Shenzhen Peacock Program under Grant No. ZX20230597, NSFC under Grant No. 62502120, the National Nature Science Foundation of China No. 62272130, Shenzhen Key Laboratory of Internet Information Collaboration,  Harbin Institute of Technology (Shenzhen), Shenzhen, 518055, China, KCXFZ20240903093006009, and SYSPG20241211173609009. It was also supported by the Major Key Project of Peng Cheng Laboratory.

\bibliography{main}

\clearpage

\input{appendix}

\end{document}

%% file: appendix.tex
\section{Appendix}

\begin{figure}[H]
    \centering
    \includegraphics[width=1\linewidth]{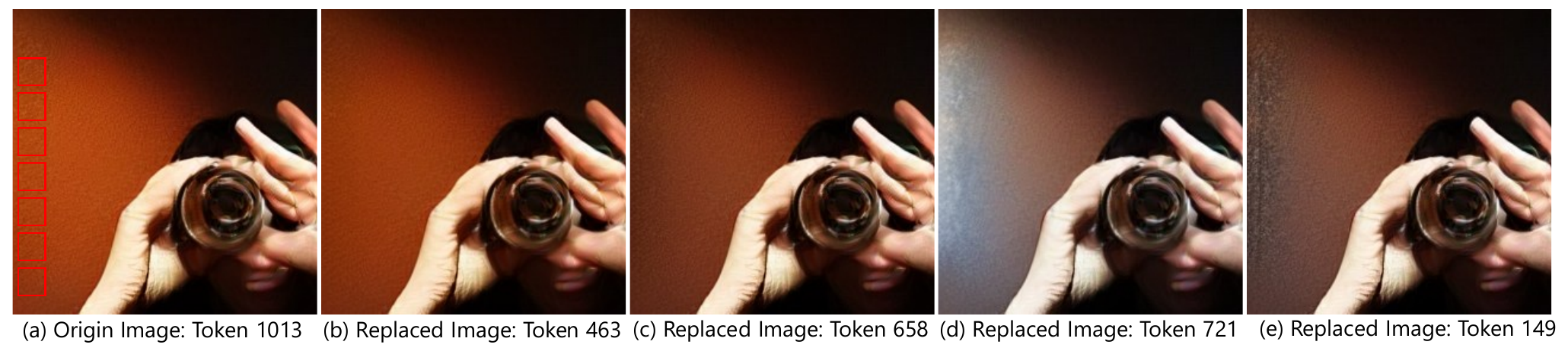}
    \caption{Visualization of semantic similarity token via reconstruction. The red box indicates the location of token (1013). We compare different versions of the same image: (a) the original reconstruction, (b), (c), (d), (e) represent replacing token (1013) with token (463), (658), (721), (149) respectively.}
    \label{fig:sim_graph_example2}
\end{figure}

\begin{figure}[H]
    \centering
    \includegraphics[width=1\linewidth]{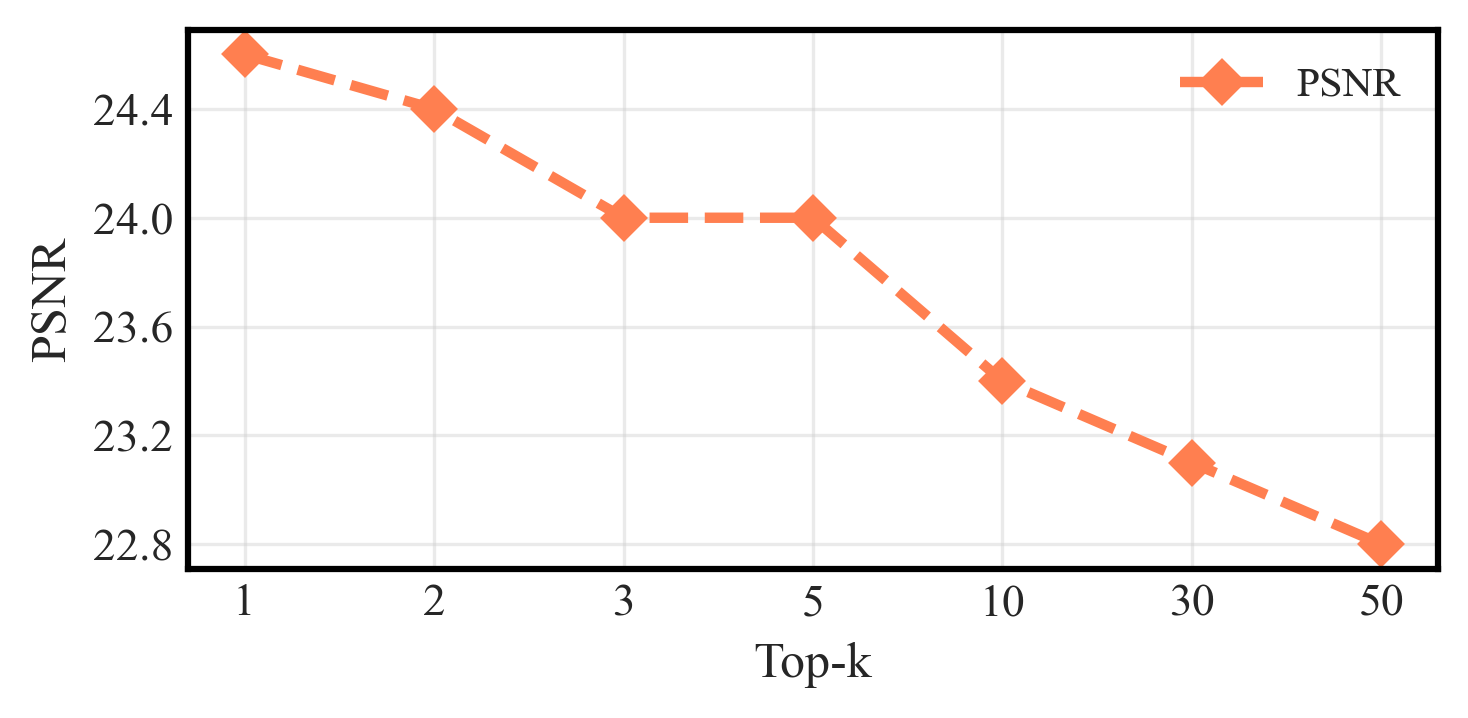}
    \caption{Results of PSNR~\cite{fardo2016formal} between origin image and top-k replacement images. We counted 1024 images.}
    \label{fig:topk2}
    \vspace{-2mm}
\end{figure}

\textbf{More Explanation for Semantic Similarity Graph}
\textit{Semantic Similarity Graph $\mathcal{G}_{s}$} indicates that two tokens express similar semantics in the context of image synthesis, which is akin to the definition of synonyms in natural language. Its advantage is that it encourages the model to capture interchangeable token patterns and improve robustness. We propose to build token similarity based on the position distribution. To further this point, as illustrated in Figure~\ref{fig:sim_graph_example2}, we replace token (1013) with its top-1 similar token (463), top-2 similar token (658), top-3 similar token (721) and least similar token (149), and input the modified token maps into the decoder for image reconstruction. The results show that replacing with token (463 and 658) yields no perceptible difference from the original image, whereas token (721 and 149) significantly degrades image quality. Moreover, we also conduct statistical analysis based on 1024 images in Figure~\ref{fig:topk2}, which shows that top-2 performs slightly below top-1 but is visually indistinguishable, while larger k degrades quality. Then, for each token, we retain the top two most similar tokens to form a directed token-to-token graph $\mathcal{G}_{s}$, which can add mild diversity and improve robustness without quality loss.


\noindent \textbf{Training and Generation Protocols}
Our training and generation configurations follow the backbone model with minor adjustments to batch sizes, learning rates, and weight regularization to obtain better performance. For the training, the detailed model configurations are shown in Table~\ref{tab:configuration}. We follow the default generation configurations of MaskGIT and AutoNAT, and set the randomize temperature of TiTok-b64 to 10.8 and the randomize temperature of TiTok-s128 to 2.9. AutoNAT and MaskGIT training and generation were conducted on 4 A100 GPUs. TiTok training was conducted on 4 A100 GPUs, and generation was conducted on 2 4090 GPUs.

\begin{table}[h]
\centering
\footnotesize
\renewcommand\tabcolsep{2.25pt}
\begin{tabular}{lcccc}
\hline
              & MaskGIT & AutoNat & TiTok-b64 & TiTok-s128 \\ \hline
Iteration     & 1500    & 1300    & 1200      & 1200       \\
Batch Size    & 100     & 300     & 500       & 500        \\
Learning Rate & 0.0001  & 0.0001  & 2.0e-06   & 2.0e-06    \\
Weight Decay  & 0       & 0       & 0         & 0.03       \\
Warmup Steps  & 1000    & 300     & 100       & 100        \\ \hline
\end{tabular}
\caption{Training configurations on different model.}
\label{tab:configuration}
\end{table}

\begin{figure*}
    \centering
    \includegraphics[width=1\linewidth]{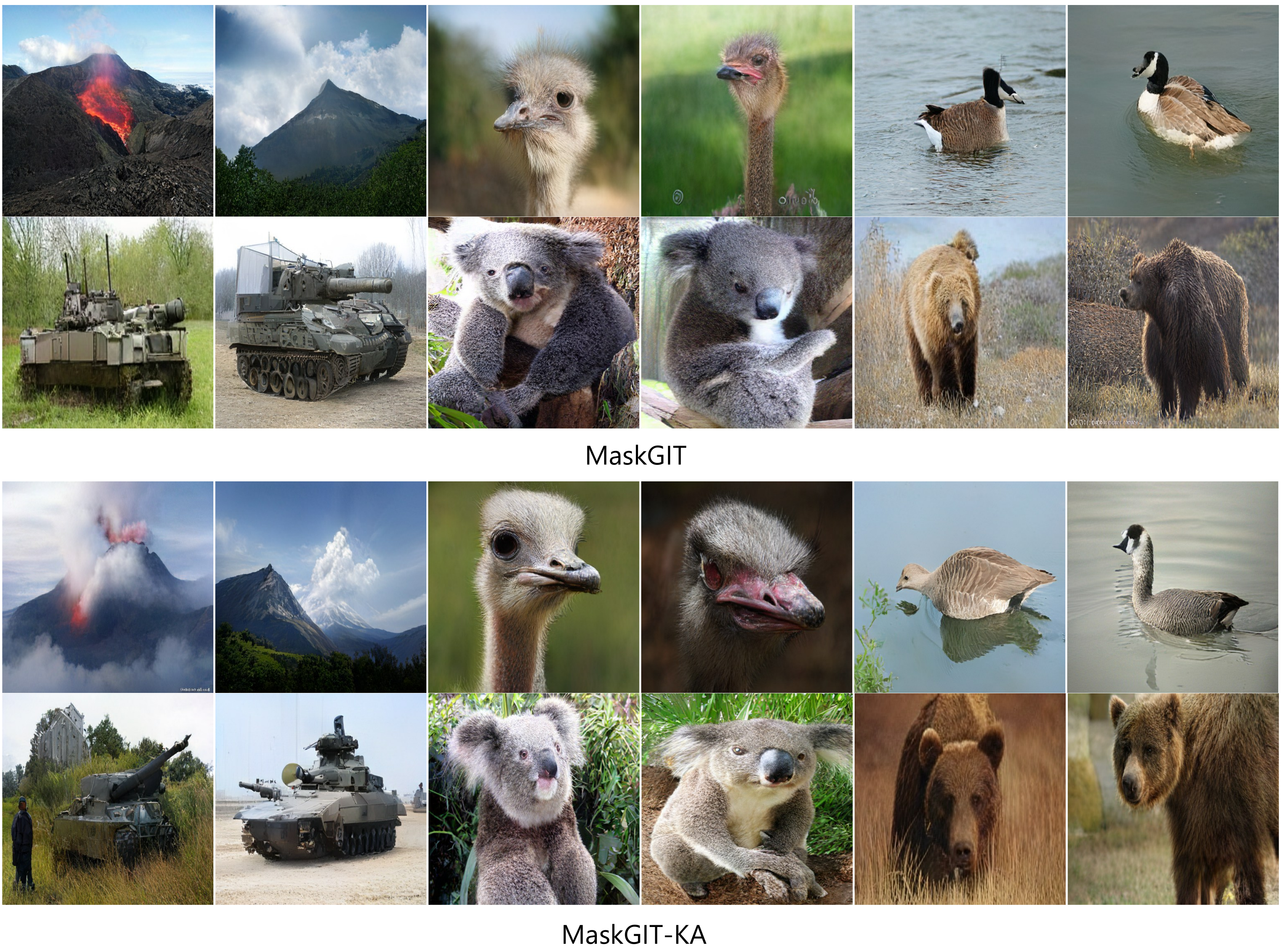}
    \caption{Visualizations for sampling images on ImageNet-512 using selected classes (980: volcano, 9: ostrich, 99: goose, 847: tank, 105: koala, 294: brown bear)}
    \label{fig:512_sampling}
\end{figure*}

\begin{figure*}
    \centering
    \includegraphics[width=1\linewidth]{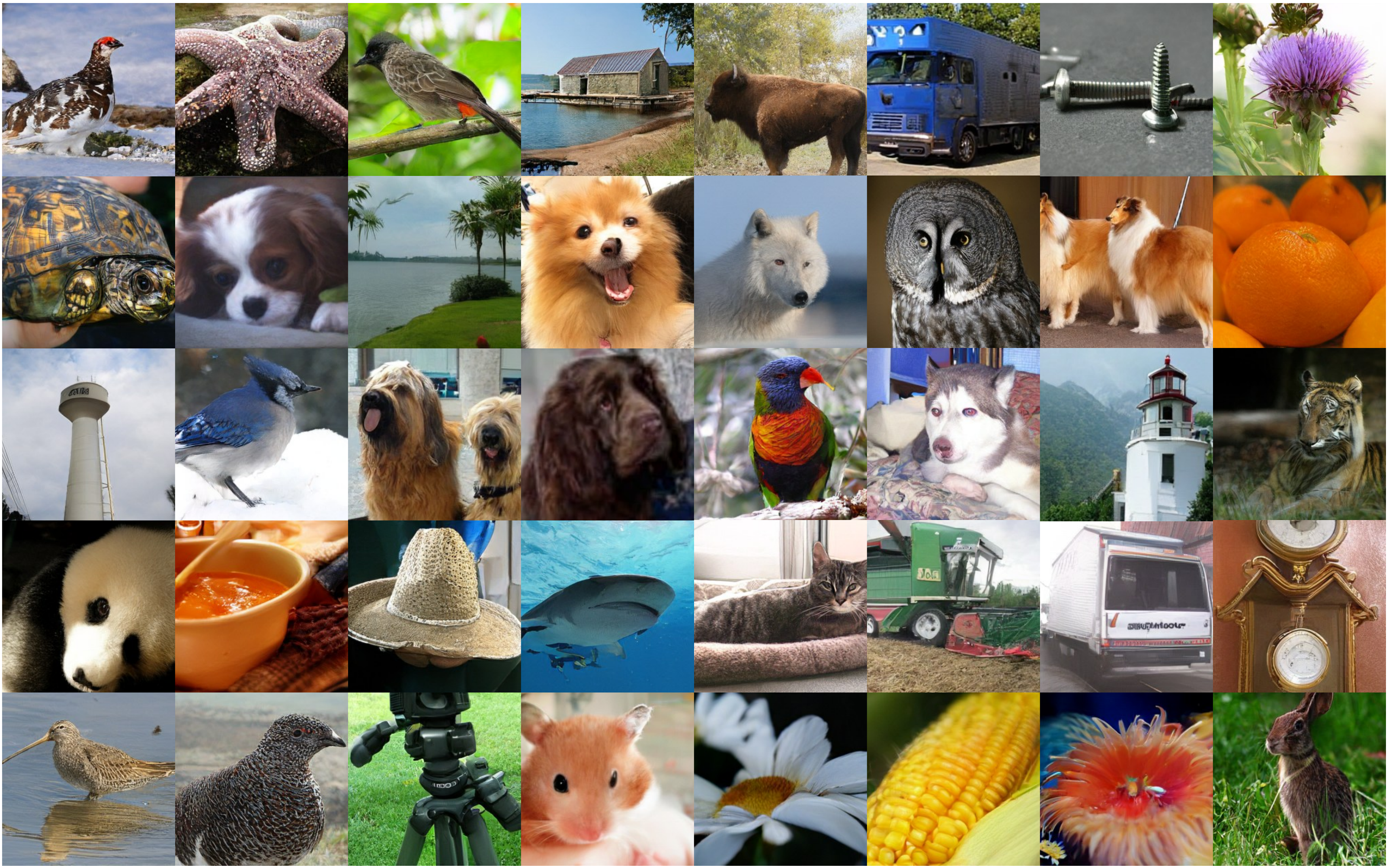}
    \caption{Visualization of generated images from \textbf{AutoNAT-KA} on ImageNet-256.}
    \label{fig:autonat}
\end{figure*}

\begin{figure*}
    \centering
    \includegraphics[width=1\linewidth]{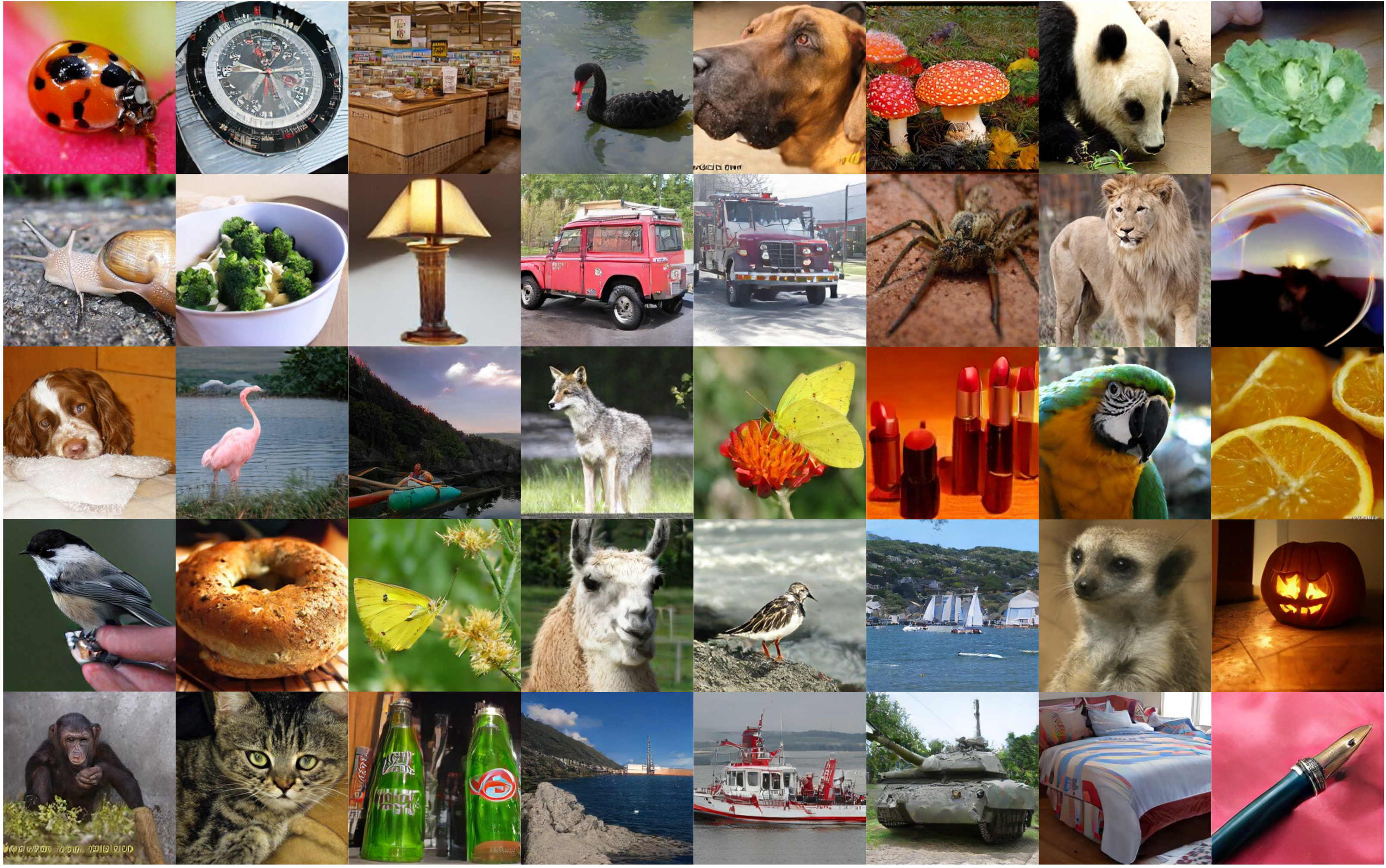}
    \caption{Visualization of generated images from \textbf{TiTok-b64-KA} on ImageNet-256.}
    \label{fig:titok64}
\end{figure*}

\begin{figure*}
    \centering
    \includegraphics[width=1\linewidth]{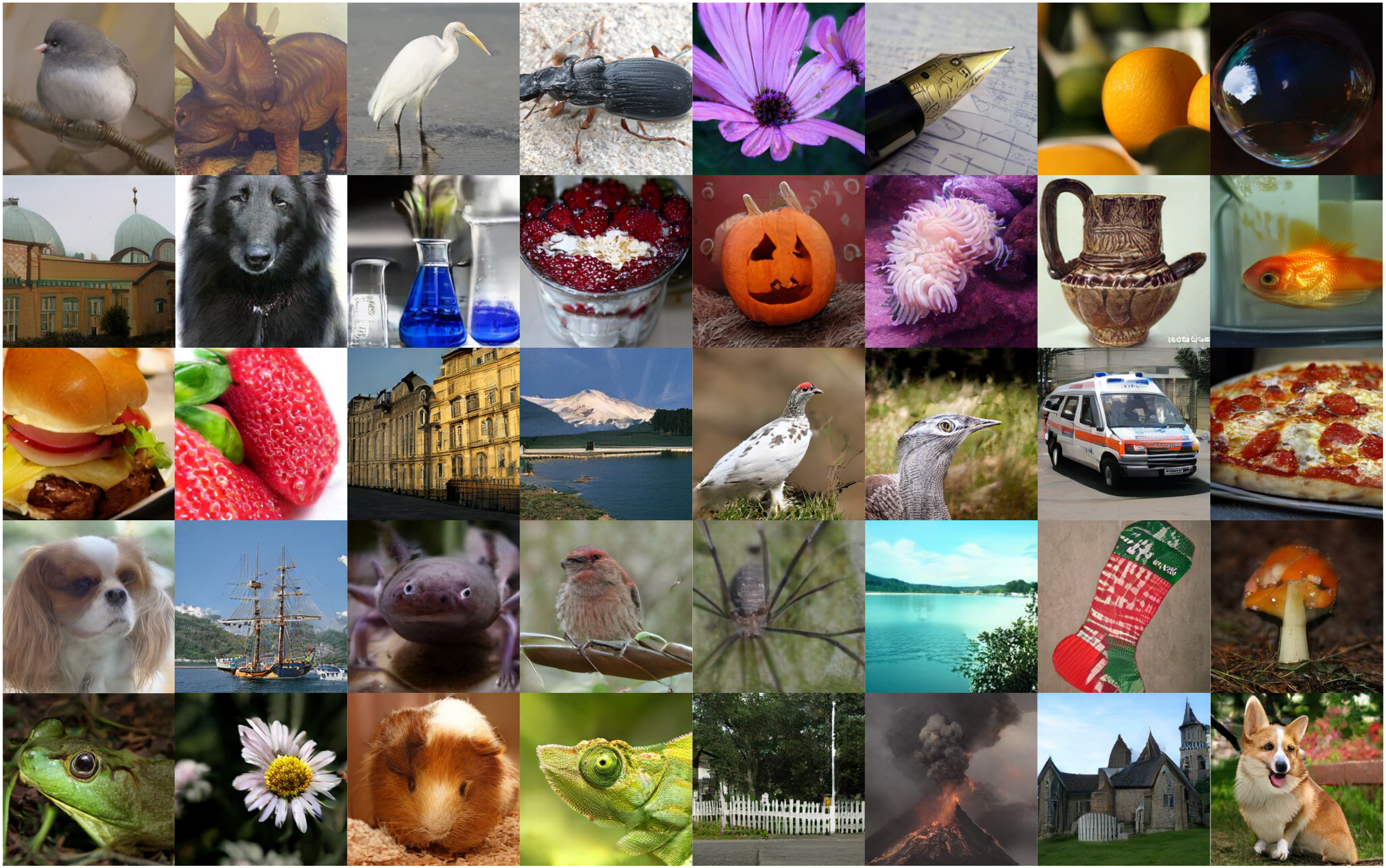}
    \caption{Visualization of generated images from \textbf{TiTok-s128-KA} on ImageNet-256.}
    \label{fig:titok128}
\end{figure*}

\noindent \textbf{More Examples}
We provide a qualitative comparison of image generation on ImageNet-512 in Figure~\ref{fig:512_sampling}. Moreover, we provide more generated images in Figure~\ref{fig:autonat} (AutoNAT-KA), \ref{fig:titok64} (TiTok-b64-KA), and \ref{fig:titok128} (TiTok-s128-KA).

